\documentclass[10pt,twocolumn,letterpaper]{article}

\usepackage{cvpr}              

\usepackage{graphicx}
\usepackage{amsmath}
\usepackage{amssymb}
\usepackage{booktabs}
\usepackage{multirow}
\usepackage[accsupp]{axessibility}

\usepackage[pagebackref,breaklinks,colorlinks]{hyperref}

\usepackage[capitalize]{cleveref}
\crefname{section}{Sec.}{Secs.}
\Crefname{section}{Section}{Sections}
\Crefname{table}{Table}{Tables}
\crefname{table}{Tab.}{Tabs.}

\def\cvprPaperID{7657} 
\def\confName{CVPR}
\def\confYear{2023}

\begin{document}

\title{Probing neural representations of scene perception in a hippocampally dependent task using artificial neural networks}

\author{
Markus Frey$^{1, 2}$\quad Christian F. Doeller$^{1,2}$\quad Caswell Barry$^{3}$
\\[1em]{\small $^1$Kavli Institute for Systems Neuroscience, NTNU, Norway\quad $^2$Max-Planck-Insitute for Human Cognitive and Brain Sciences, Germany\quad} \\ {\small $^3$Cell \& Developmental Biology, UCL, United Kingdom}\vspace{-1em}}

\maketitle


\begin{abstract}
Deep artificial neural networks (DNNs) trained through backpropagation provide effective models of the mammalian visual system, accurately capturing the hierarchy of neural responses through primary visual cortex to inferior temporal cortex (IT)\cite{yamins2016using, zhuang2021unsupervised}. However, the ability of these networks to explain representations in higher cortical areas is relatively lacking and considerably less well researched. For example, DNNs have been less successful as a model of the egocentric to allocentric transformation embodied by circuits in retrosplenial and posterior parietal cortex. We describe a novel scene perception benchmark inspired by a hippocampal dependent task, designed to probe the ability of DNNs to transform scenes viewed from different egocentric perspectives. Using a network architecture inspired by the connectivity between temporal lobe structures and the hippocampus, we demonstrate that DNNs trained using a triplet loss can learn this task. Moreover, by enforcing a factorized latent space, we can split information propagation into "what" and "where" pathways, which we use to reconstruct the input. This allows us to beat the state-of-the-art for unsupervised object segmentation on the CATER and MOVi-A,B,C benchmarks. 
\end{abstract}

\section{Introduction}
\label{sec:intro}

Recently, it has been shown that neural networks trained with large datasets can produce coherent scene understanding and are capable of synthesizing novel views \cite{eslami2018neural, kabra2021simone}. These models are trained on egocentric (self-centred) sensory input and can construct allocentric (world-centred) responses. In animals, this transformation is governed by structures along the hierarchy from the visual cortex to the hippocampal formation, an important model system related to navigation and memory \cite{hubel_receptive_1962, okeefe_hippocampus_1971}. Notably, the hippocampus is a necessary component of the network supporting memory and perception of places and events and is one of the first brain regions compromised during the progression of Alzheimer's disease (AD) \cite{bird2008hippocampus, vsimic1997volume}. However, experimental knowledge regarding the interplay across multiple interacting brain regions is limited and new computational models are needed to better explain the single-cell responses across the whole transformation circuit.

Here, we developed a scene recognition model to better understand the intrinsic computations governing the transformation from egocentric to allocentric reference frames, which controls successful view synthesis in humans and other animals. For this, we developed a novel hippocampally dependent task, inspired by the 4-Mountains-Test \cite{hartley2007hippocampus}, which is used in clinics to predict early-onset Alzheimer's disease \cite{wood2016allocentric}. We tested this task by creating a biologically realistic model inspired by recent work in scene perception, in which scenes need to be re-imagined from several different viewpoints. 

The main contributions of our paper are the following: 
\begin{itemize}
    \item We introduce and open-source the allocentric scene perception (ASP) benchmark for training view synthesis models based on a hippocampally dependent task which is frequently used to predict AD.
    \item We show that a biologically realistic neural network model trained using a triplet loss can accurately distinguish between hundreds of scenes across many different viewpoints and that it can disentangle object information from location information when using a factorized latent space. 
    \item Lastly, we show that by using a reconstruction loss combined with a pixel-wise decoder we can perform unsupervised object segmentation, outperforming the state-of-the-art models on the CATER, MOVi-A,B,C benchmarks.
\end{itemize}

\begin{figure*}
  \centering
  \includegraphics[width=\linewidth]{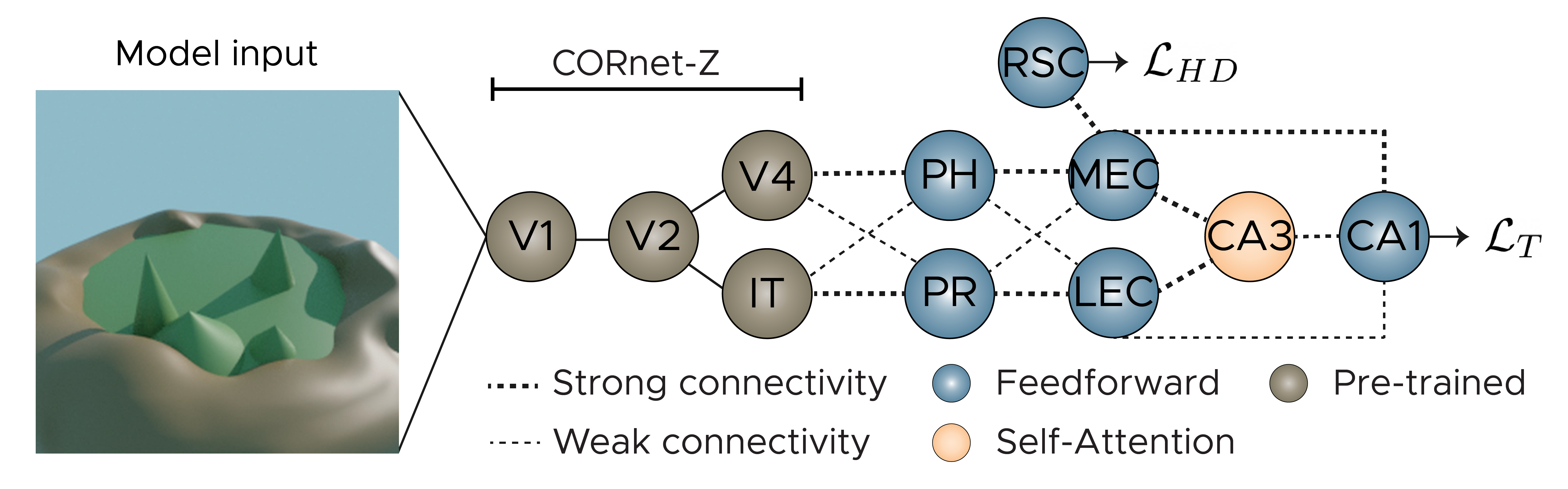}
  \caption{\textbf{Task design and model architecture} \\ We adapt the 4-Mountains-Test \cite{hartley2007hippocampus}  using a simplified task design with one to four objects with circular symmetry and a distal landmark, depicting a mountain range. We then design a biologically inspired model architecture, for which we take visual cortex responses from pre-trained CORnet-Z \cite{kubilius2018cornet} and feed them through perirhinal (PR) and parahippocampal (PH) cortices. The retrosplenial cortex uses an auxiliary loss to decode head direction. Medial entorhinal (MEC) and lateral entorhinal cortex (LEC) use a disentangled latent space to separate object from location information. Both are integrated within the hippocampus, with CA3 using a self-attention layer across both time and space. }
  \label{fig1}
\end{figure*}

\section{Related work}
\subsection{Related work in neuroscience}
The study of scene perception in neuroscience was first systematically explored in behavioural experiments in 1972, by showing for the first time scenes in real-world contexts to human participants \cite{biederman1972perceiving}. These outdoor scenes were either shown intact or scrambled into six randomly arranged pieces. It was shown, that the correct identification of cued objects was drastically lower when the scenes were scrambled, indicating that the various parts of a scene are perceived as a whole. 

Neural recordings as a response to scenes were first reported in fMRI experiments that showed that the parahippocampal place area (PPA) is involved in perceiving the visual environment, being more active for outdoor scenes than single objects \cite{epstein1998cortical}. Moreover, activity is reduced when scrambling the picture into random pieces, indicating that neurons in PPA are sensitive to the structure of the entire scene \cite{epstein2008parahippocampal}. A second scene-selective region was shown to be also active during mental imagination of scenes, subsequently labelled as the retrosplenial complex (RSC). The RSC also acts as a hub to integrate sensory, motor and visual information and is crucial for the transformation from egocentric to allocentric reference frames \cite{nau2020behavior, clark2018retrosplenial}. 

Computational modelling has suggested that the transformation in RSC is governed by head direction cells \cite{byrne2007remembering}, which encode the world-centred facing direction of the head of the animal in the azimuthal plane \cite{ranck1985head}. When egocentric information from the parietal cortex reaches RSC, the head direction signal aligns these cells to a common reference, creating complementary allocentric cell types. These include object vector cells in the medial entorhinal cortex \cite{hoydal2019object}, which are active at spatially confined objects and boundary vector cells (BVCs) in the subiculum \cite{lever2009boundary} which encode boundary information in their neuronal activity. The combined activity of several BVCs giving rise to allocentric place cells in the hippocampus \cite{bicanski2018neural, byrne2007remembering}. 

In recent years, normative machine-learning approaches have investigated visual information processing in an egocentric reference frame while translating it into an allocentric representation for downstream task performance. Several types of these models have been explored, most notably the Tolman-Eichenbaum-Machine (TEM) \cite{whittington2020tolman} and the Spatial Memory Pipeline (SMP) \cite{uria2020spatial}. These normative models have shown the emergence of allocentric spatial cells in two separate tasks, using a similar objective function: predicting the next sensory observation. However, both of these models have not been tested on their abilities to generate images from novel viewpoints, with TEM only taking in abstract sensory observations and SMP being employed in a reinforcement-learning (RL) environment using a spatial navigation task. 

\subsection{Related work in computer science}
Beyond TEM and SMP, specially designed scene perception models have tackled the problem of novel view synthesis in several distinct ways. For example, in robotics, SLAM (Simultaneous localization and mapping) algorithms have been used extensively to represent scenes and navigate within them, but mainly in a supervised setting on partially occluded scenes and with a focus on the navigational abilities of robots, which are endowed with additional sensors \cite{cadena2016past}.  

Neural radiance field (NERF) networks \cite{mildenhall2020nerf, martin2021nerf, tancik2020fourier} try to mimic the image synthesis based on real-world physics, namely the way light is reflected from certain materials and how these rays end up in the camera sensor. For this, the neural network must infer all the scene's physical properties. The actual reconstruction of an image is done analytically by a traditional rendering engine, which is typically not changed during training \cite{tewari2022advances}. The advantage of these rendering methods lies in the fact that arbitrary resolutions can be achieved. This is possible as it is not using image or voxel space, but a continuous neural representation, where coordinates are mapped through a neural network to their corresponding value - representing colour, occupancy or material properties \cite{pumarola2021d, martin2021nerf}. However, a common shortfall of neural rendering methods is the use of separate neural networks for each scene, making it hard to gauge the generalization abilities beyond its training scenes. 

Traditional scene decomposition methods beyond NERFs are able to generalize across many scenes, while still relying on pixel-level information. Most commonly, the model takes several snapshots of a scene from different viewpoints which are joined in a latent space. To produce new views, the latent space is conditioned with a new set of camera coordinates, reconstructing a novel viewpoint \cite{eslami2018neural}. Recent models have focused on architectures using a slot structure which disentangles objects within a scene from each other using self-attention \cite{locatello2020object, greff2020binding, kabra2021simone}. This allows the model to learn a fully unsupervised factorized latent space, which is used to synthesize novel viewpoints and scenes with different compositionality \cite{kabra2021simone}. 

\section{Task design}

\begin{figure}
  \centering
  \includegraphics[width=\linewidth]{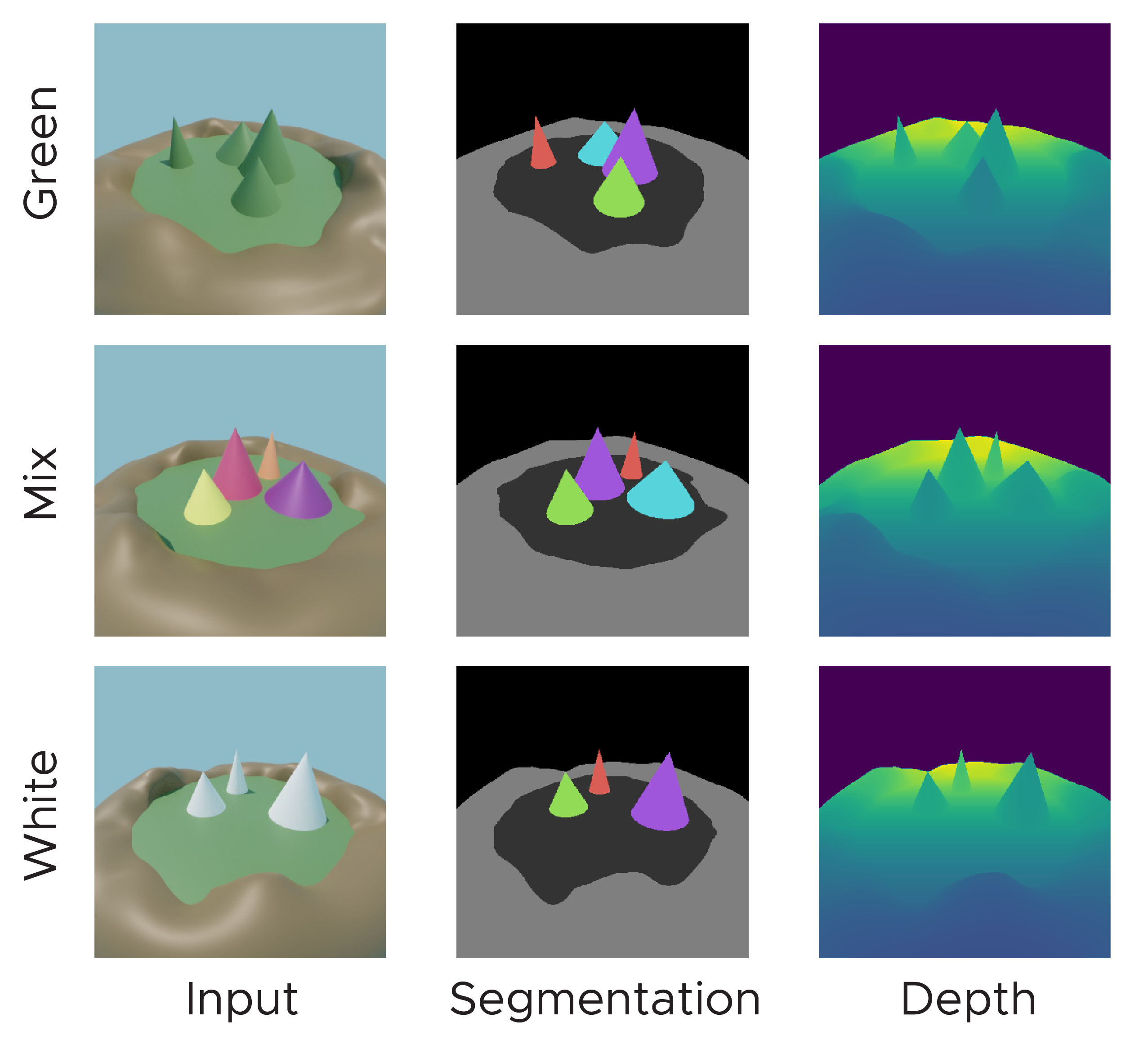}
  \caption{\textbf{Variations of task design} \\ We render the task in six different versions (three shown here), varying both the number of objects within the scenes from 1 to 4 as well as object colour. We provide segmentation masks, including objects, backgrounds, and depth maps. We also render each task without a distal landmark; see Figure \ref{fig_s2}.}
  \label{fig5}
\end{figure}

None of the above-mentioned biologically inspired models is equipped to deal with randomly sampled egocentric sensory observations. Therefore we built a model inspired by novel view synthesis tasks while considering the particularities of the hippocampally dependent 4-Mountains-Test \cite{hartley2007hippocampus, chan_listen_2016}. This test is used in the clinic and requires allocentric topographical processing for successful task performance, thus is sensitive to hippocampal damage including that which accumulates during the early stages of AD. The participant first views an image of a scene with four mountains in it and after a two-second delay is asked to match the same-scene image out of four images (three distractors and one target image, see Figure \ref{fig_s1} for the same test using our adapted design). Most importantly, the correct image is the same scene, seen from a different viewpoint. In contrast, the distractor images show scenes where the objects are located in different allocentric configurations compared to the original scene. 

We simplified the task into its core components by rendering four objects with circular symmetry and a global reference frame given by the surrounding landscape. Novel scenes were rendered by randomly changing the world-centred location of each object, thereby changing the relative distances of the objects to themselves and the boundary. We acquire egocentric sensory observations by rendering the scene across different viewpoints, varying both azimuth and elevation (Figure \ref{fig5}). We also render views from the 'inside' of the environment, simulating the views of an agent navigating the scene. We render six different versions of this task, by varying both the distal landmark $(with | without)$ and the colours of the objects $(mix | green | white)$, while randomly varying the number of objects within the scene from 1 to 4, with 10\% of the scenes having one object, 20\% two, 30\% three, and 40\% containing four objects. We sampled different colours, as traditional scene perception models struggle to differentiate objects with the same colour or the floor's colour. During rendering, we also store segmentation masks and depth profiles for each rendered viewpoint (Figure \ref{fig5} \& Figure \ref{fig_s2}). Moreover, we store positional information for analysis of neural activity with respect to different reference frames. This includes the positions of the objects in an egocentric reference frame (position 'on the screen'), angular information such as azimuth and elevation and the allocentric positions of each object within the environment (Figure \ref{fig3}). 

\section{Model architecture}

The overall architecture of the model follows known biological connectivity between visual cortices and the hippocampal formation \cite{bird2008hippocampus} (Figure \ref{fig1}). To simulate the responses of the visual cortex, we use a pre-trained convolutional neural network which accurately predicts neural responses from the macaque visual system \cite{kubilius2018cornet, schrimpf2020brain} (CORNet-Z). This ensures that visual cortex responses do not overfit the objects, colours and backgrounds we use but generalize onto natural images. We extract activations from visual cortex area 4 (V4) and inferior temporal (IT) cortex for each rendered viewpoint. 

The information from V4 and IT is then routed through perirhinal (PR) and parahippocampal (PH) cortices using either weak or strong connectivity. We implement the strong connections using a feedforward layer with non-linearity between two areas, and the weak connections by adding a residual connection consisting of a linear feedforward layer that uses the summed input of all previous activations. Both areas receive information mainly from the ventral visual processing stream, with PR being crucial for the representation of objects ('what'), while PH predominantly processes visuospatial information \cite{bird2008hippocampus}. We implement this split of information by averaging the visual cortex representation either across the temporal or spatial axis, which also enforces disentangled latent representations in subsequent layers, namely the medial (MEC) and lateral (LEC) entorhinal cortex. MEC receives input which is averaged across time, therefore being trained to keep spatial information, while LEC receives input averaged across the spatial dimension, thereby retaining temporal information. 
We implement the hippocampus consisting of CA3 and CA1, with the former integrating information using a self-attention layer across different snapshots of the same scene and the latter integrating both temporal and spatial information using the outer product, which is fed through a simple feedforward layer, similar to how TEM integrates information in the hippocampal formation \cite{whittington2021relating}.

To reconstruct the input across novel views, we use a pixel-wise decoder similar to the one used in SIMONe \cite{kabra2021simone}. For the feedback connections, we use the CA1 layer which is split into temporal (LEC) and spatial (MEC) information exactly as the forward pass splits the visual information at the level of the PR/PH layers. We then sample from LEC and MEC for each object and pixel and combine them with periodic positional encodings. These are fed through a 5-layer MLP decoder using 128 units each, decoding RGB values for each pixel and object which are combined with alpha masks to produce the full reconstructed image (Figure \ref{fig_s4}).

\section{Model optimization}
During model training, the model receives egocentric sensory observations which are transformed into allocentric representations in the hippocampal formation. The model is trained by minimizing either a triplet loss on this hippocampal latent space or an L2-reconstruction loss in pixel space if the model is tasked to reconstruct the image:
\begin{equation}
    \mathcal{L}_{a,p,n} = \sum_{i}\max\left(\sqrt{(x_{i}^{a} - x_{i}^{p})^2} - \sqrt{(x_{i}^{a} - x_{i}^{n})^2} + \alpha, 0\right)
\end{equation}
\begin{equation}
    \mathcal{L}_{MSE} = \frac{1}{n}\sum_{n}{(x_{i} - \hat{x}_{i})^2}
\end{equation}
where $x^{a,p,n}$ denotes the anchor, positive or negative representation from a given layer. If the image layer is used, these correspond to the original image in pixel space. The L2-reconstruction loss is calculated between the predicted reconstruction $\hat{x}_{i}$ and the original image $x_{i}$ on the full-resolution image. The final loss is summed across the width, height and timesteps and averaged across batches.

The predictive objective function used in previous models \cite{whittington2020tolman, uria2020spatial} is similar to the reconstruction loss in our model as the latent space enforces the reconstruction of scenes from different viewpoints, which can be understood as a prediction of not only the next observation but all possible observations. Note that many recent models capable of novel view synthesis use variational inference \cite{kabra2021simone, burgess2019monet}, for which it is unclear how individual distribution statistics would be sampled in biological tissue. We, therefore, do not sample from the distribution and only use the mean, similar to how TEM constructs its latent space \cite{whittington2020tolman}.

\section{Results}

\begin{figure}
  \centering
  \includegraphics[width=\linewidth]{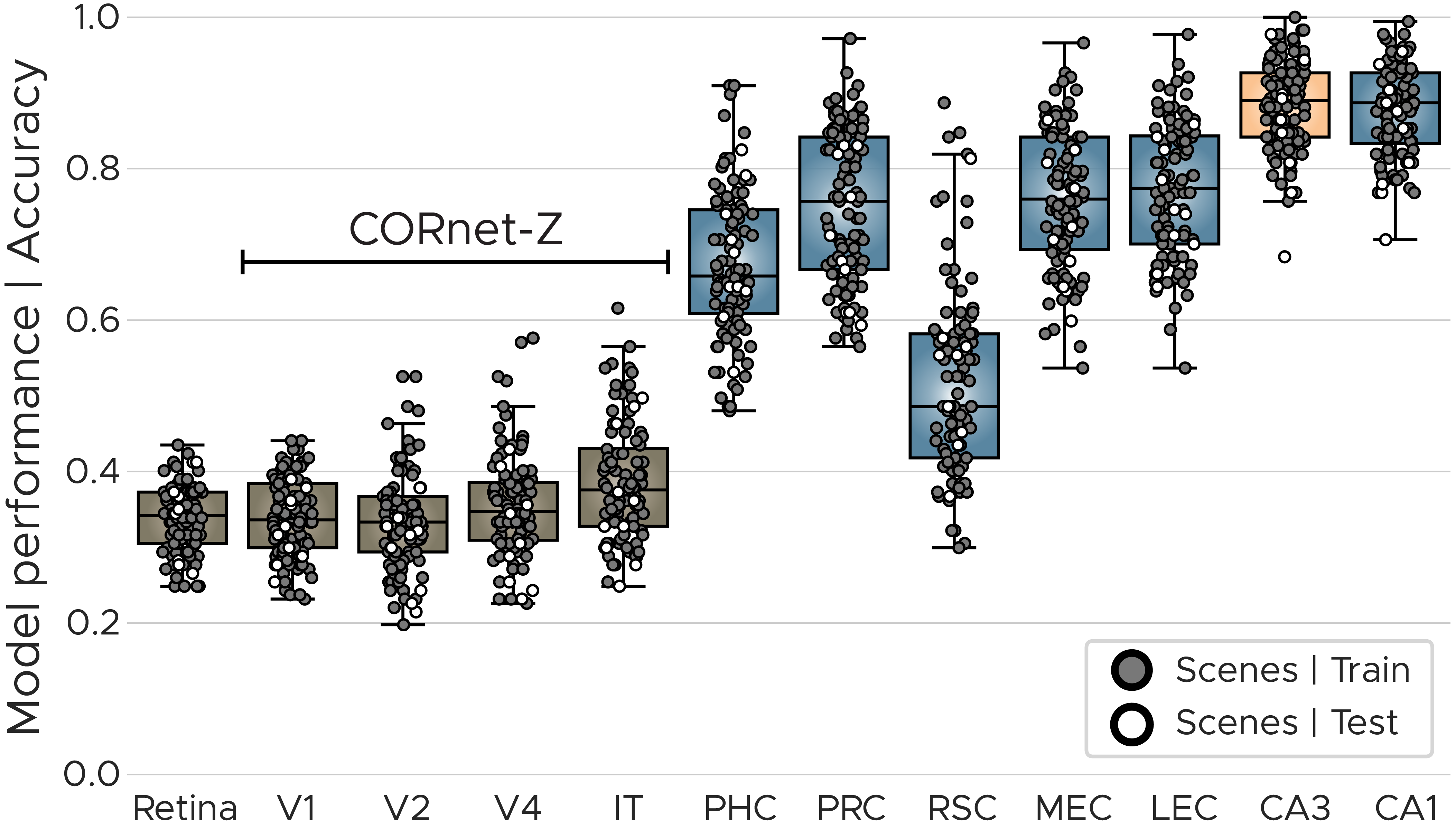}
  \caption{\textbf{Model performance on scene perception task.} Model performance across layers. For each layer, we calculate the accuracy by sampling triplets and quantifying the number of correctly classified same-scene images. Scenes used for training are depicted as grey circles and test scenes as white circles. Most world-centred responses are measured in late layers, indicating that hippocampal responses are similar for images from the same scene from different viewpoints. }
  \label{fig2}
\end{figure}

\subsection{Performance on adapted 4-Mountains-Test}

We first train our scene perception model to separate between different scenes, closely linked to the original task in the 4-Mountains-Test (Figure \ref{fig_s1}). We use a triplet loss in which we sample an anchor and a positive image from the same scene, together with a negative image from a different scene. This contrastive loss function allows us to disentangle the hippocampal representation between different scenes maximally. Model performance is evaluated by randomly sampling triplets (anchor, positive, negative) from the whole dataset, calculating a cross-correlation matrix and using the smallest off-diagonal value as the correct image. If a layer is able to distinguish between scenes, the correlation between the same-scene pair seen from different viewpoints should be high, while the correlation between different-scene pairs should be low.

This performance measure allows us to evaluate accuracy across all model layers. We observe that the pre-trained layers and the image itself show performance levels around chance, with IT performing best (Chance: 33\%, IT: 39\%, Figure \ref{fig2}). This means that even though IT is thought to contain object-specific information, it lacks crucial information to differentiate between scenes containing the same objects but in different allocentric positions. Nevertheless, these scenes can be distinguished by late layers in our model, with CA3 and CA1 showing performance close to 90\% (CA3 89\%, CA1 88\%), indicating that these layers construct a world-centred representation of the environment, which is needed for separating the scenes in the adapted 4-Mountains-Test.


\subsection{Neural representations within network layers}

To quantify the amount of allocentric information contained in each trained layer, we calculate an allocentricity measure. We define allocentricity as the coefficient of variation of the activation of each artificial neuron across several images of the same scene. A neuron that fires similarly across images of the same scene from a different viewpoint has a high allocentric score, while a neuron with a high variance in its activations for the same stimuli has a low allocentric score. We observe that the hippocampal layers show the highest allocentricity score (PH, -5.5$\pm$0.03; PR, -2.6$\pm$0.05; CA3, 5.0$\pm$2.2; CA1, 3.5 $\pm$ 1.7), indicating that these layers have learned to be active for the same scene across different viewpoints, i.e. have formed an allocentric representation of the environment (Figure \ref{fig_s3}). 

Having established a differentiation across layers between egocentric and allocentric information, we sought to investigate which scene properties are represented in each layer. We use a linear readout to investigate the separation of scene properties into low-level features like colours or the position on the screen, mid-level elements like object size and high-level features like allocentric position and scene identity \cite{epstein2019scene}. We observe a trend toward later layers incorporating more high-level information. However, the overall structure is less clear than previously reported in the visual cortex \cite{yamins2014performance, khaligh2014deep, gucclu2015deep}. Information regarding the position of objects in allocentric coordinates can only be effectively read out from hippocampal layer CA3, while egocentric information - object's position in screen coordinates - is reduced drastically in the layers beyond MEC, with CA1 seemingly only retaining allocentric information (Figure \ref{fig_s3}).

\begin{figure}
  \centering
  \includegraphics[width=\linewidth]{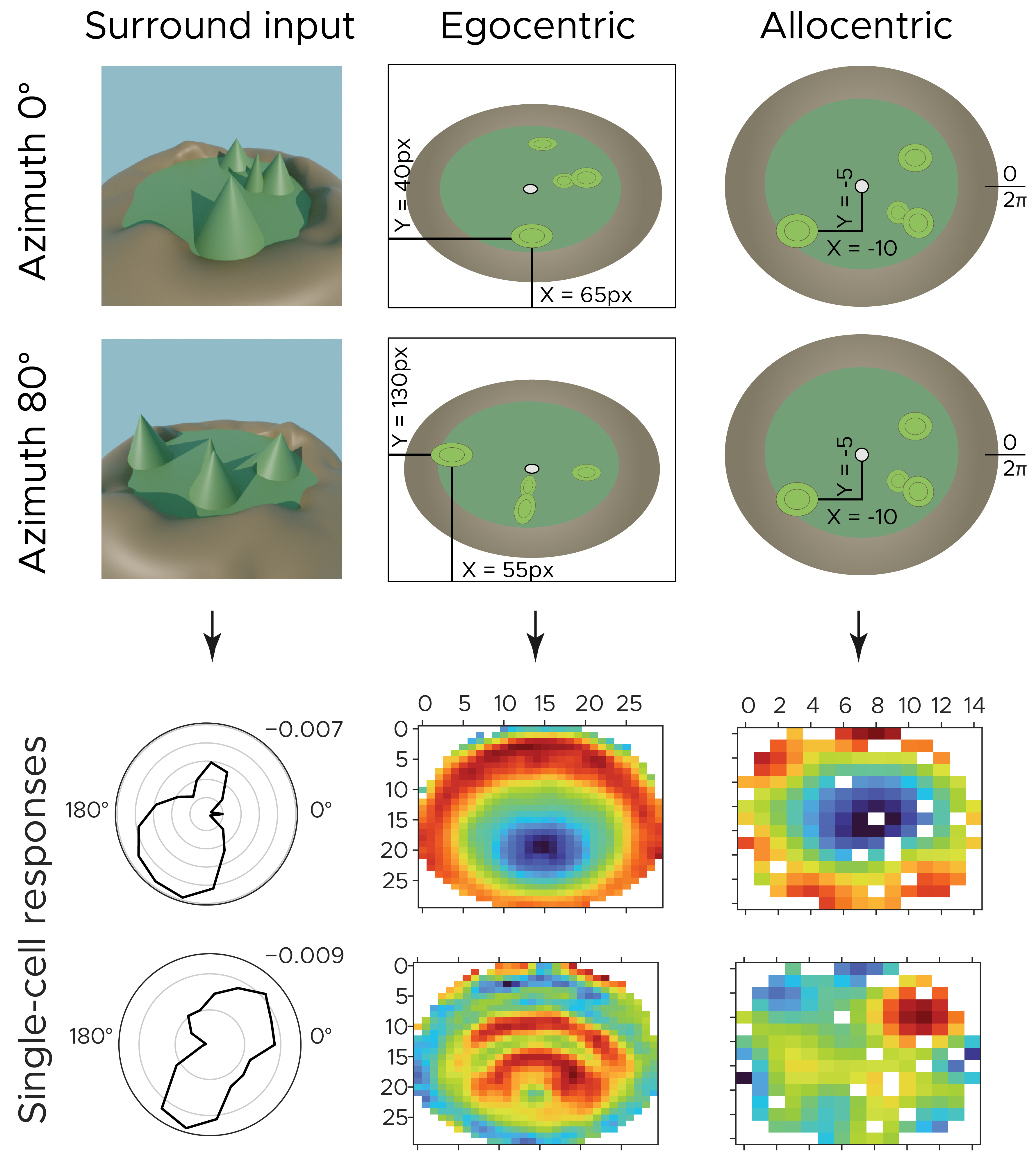}
  \caption{\textbf{Single-cell representations across different reference frames.} (Top) Illustration of
egocentric and allocentric schemas. The left side shows two snapshots from the same scene using different viewpoints. The panels to the right depict the egocentric and allocentric schema for the respective snapshot. Black lines indicate screen coordinates for the same object in the egocentric view and world coordinates in the allocentric view. Note that by definition the allocentric, world-centred schema is the same for both snapshots. (Bottom) Example representations from neurons within the network, across each reference frame, showing activity of two neurons for azimuth, egocentric object position and allocentric object position. }
  \label{fig3}
\end{figure}

We next sought to investigate individual neural responses to obtain a fine-grained understanding of the computations being performed within each layer and, importantly, the ways in which they relate to known biological data. For this purpose, we visualize the activity of a subset of neurons within each layer with regard to different reference frames (Figure \ref{fig3}). We use the egocentric reference frame for pixel coordinates and the allocentric reference frame for objects or locations in the environment. To visualize angular allocentric responses, we visualize the activity using polar coordinates. We observe allocentric boundary and place-like activity in the CA1 layers as a function of the position of an object (Figure \ref{fig3}). This indicates that single neurons within the model layer are highly activated whenever an object is close to the boundary or occupies a certain allocentric position within the environment, similar to the boundary and place-like activity observed in biological organisms \cite{lever2009boundary, okeefe_hippocampus_1971}. These spatial responses also show similar mechanisms to biological cells, as they tend to remap across different scenes (\ref{fig_s5}. We observe these responses as a result when using only the triplet loss on the CA1 layer. As we describe further below, we can reconstruct and segment the image into distinct objects using an additional reconstruction loss.



\subsection{Reconstructing the input through feedback connections}

Having established that the model is able to discern between novel scenes from arbitrary viewpoints, we next explored its ability to reconstruct the scene across these viewpoints. For this, we added an additional reconstruction loss in pixel space (Figure \ref{fig_s4}). Reconstructing the input image is more challenging than just differentiating between scenes, as complete scene information has to be retained across layers or reinstated from a latent representation. Therefore, we used a factorized latent space to sample objects and frame information for each individual pixel and time point, similar to recent scene perception models \cite{kabra2021simone}. It is assumed that mental imagination (and similarly novel view synthesis) is guided by a viewpoint-changing signal likely provided by grid cells in the medial entorhinal cortex, which is combined with object information in the lateral entorhinal cortex and is then further disentangled into egocentric information inside the retrosplenial cortex, which together with visual cortex establishes the mental image \cite{bicanski2018neural}. 

We first test the reconstruction loss across our six variations of the task, in which we varied object colour and background information (Figure \ref{fig_s2}). We observe a difference in the segmentation performance of the model depending on the object colours used, with the green and white colours performing worse than the mixed objects, likely because it is harder to differentiate between objects of the same colour (Table \ref{tab2}). This has also been noticed in traditional scene decomposition models, which struggle to disentangle objects of the same colour and objects having a similar colour to the background \cite{eslami2018neural}. Interestingly, if we do not enforce the disentanglement in CA1 via the triplet loss, we can still see a differentiation between scenes in the late layers of the model by just training on the reconstruction loss. 

\begin{table}
\centering
\begin{tabular}{llll}
                        &       & Distal landmark & No landmark  \\ 
\hline
\multirow{3}{*}{MSE}    & Green & \textbf{27.912 {\scriptsize $\pm$ 1.289}}              & 46.411 {\scriptsize $\pm$ 2.777}           \\
                        & Mix   & 45.908 {\scriptsize $\pm$ 4.405}             & 60.159 {\scriptsize $\pm$ 1.923}           \\
                        & White & 52.713 {\scriptsize $\pm$ 4.119}              & \textbf{19.167 {\scriptsize $\pm$ 0.575}}           \\ 
\hline
\multirow{3}{*}{FG-ARI} & Green & 0.137 {\scriptsize $\pm$ 0.122}              & 0.046 {\scriptsize $\pm$ 0.010}           \\
                        & Mix   & \textbf{0.207 {\scriptsize $\pm$ 0.148}}              & \textbf{0.297 {\scriptsize $\pm$ 0.119}}           \\
                        & White & 0.086 {\scriptsize $\pm$ 0.035}             & 0.140 {\scriptsize $\pm$ 0.023}           \\ 
\hline
\multirow{3}{*}{ARI}    & Green & 0.122 {\scriptsize $\pm$ 0.063}              & \textbf{0.067 {\scriptsize $\pm$ 0.066}}           \\
                        & Mix   & \textbf{0.383 {\scriptsize $\pm$ 0.122}}              & 0.016 {\scriptsize $\pm$ 0.016}          \\
                        & White & 0.136 {\scriptsize $\pm$ 0.089}              & 0.039 {\scriptsize $\pm$ 0.033}           \\
\hline
\end{tabular}
\medskip 
\caption{\textbf{Reconstruction and unsupervised segmentation performance on ASP} \\
We compare the mean-squared error for reconstructing the input images (MSE), the foreground adjusted rand index (FG-ARI) and the full adjusted rand index (ARI) across all six variations of our dataset. We use random seeds to report the mean and standard deviation across five model runs. The lowest MSE values and highest ARI values are displayed in bold.}
\label{tab2}
\end{table}


\begin{table}[tp]
\centering
\scalebox{0.80}{
\begin{tabular}{l c c c c c}
       & MONet & SIMONe & SAVi & Ours \\ 
\toprule
CATER &  0.412 {\scriptsize $\pm$ 0.012}    &   0.918 {\scriptsize $\pm$ 0.036}   &  0.928 {\scriptsize $\pm$ 0.008}  &  \textbf{0.939 {\scriptsize $\pm$ 0.013}}\\ 
MOVi-A    &    -   &     0.618 {\scriptsize $\pm$ 0.200}          &   0.820 {\scriptsize $\pm$ 0.030}     &  \textbf{0.790 {\scriptsize $\pm$ 0.017}}  \\
MOVi-B    &    -   &  0.307 {\scriptsize $\pm$ 0.330}   &   0.615 {\scriptsize $\pm$ 0.030}   &   \textbf{0.460 {\scriptsize $\pm$ 0.033}}  \\
MOVi-C    &    -  &      0.198 {\scriptsize $\pm$ 0.005}          &   0.470 {\scriptsize $\pm$ 0.030}       &   \textbf{0.318 {\scriptsize $\pm$ 0.009}}  \\
\bottomrule
\end{tabular}}
\medskip 
\caption{\textbf{Segmentation performance across unsupervised models on CATER and MOVi.} \\
We compare the foreground adjusted rand index (FG-ARI) across different unsupervised models, quantifying how well the model segmentation matches the real masks. Note that MONet is a static-frame model which predicts segmentation for each frame separately, likely causing the model to fail to track objects stably. SAVi is not a fully unsupervised model, using optical flow as a supervision signal. Baseline scores for MONet, S-IODINE and SIMONe were taken from \cite{kabra2021simone}, SAVi score was taken from \cite{kipf2021conditional}. To obtain the final model performance, we first train a model for 200000 steps with a fixed learning rate and subsequently present the efficacy of five models initialized with these weights and trained utilizing an annealing learning rate.}

\label{tab1}
\end{table}
\begin{figure*}
  \centering
  \includegraphics[width=\linewidth]{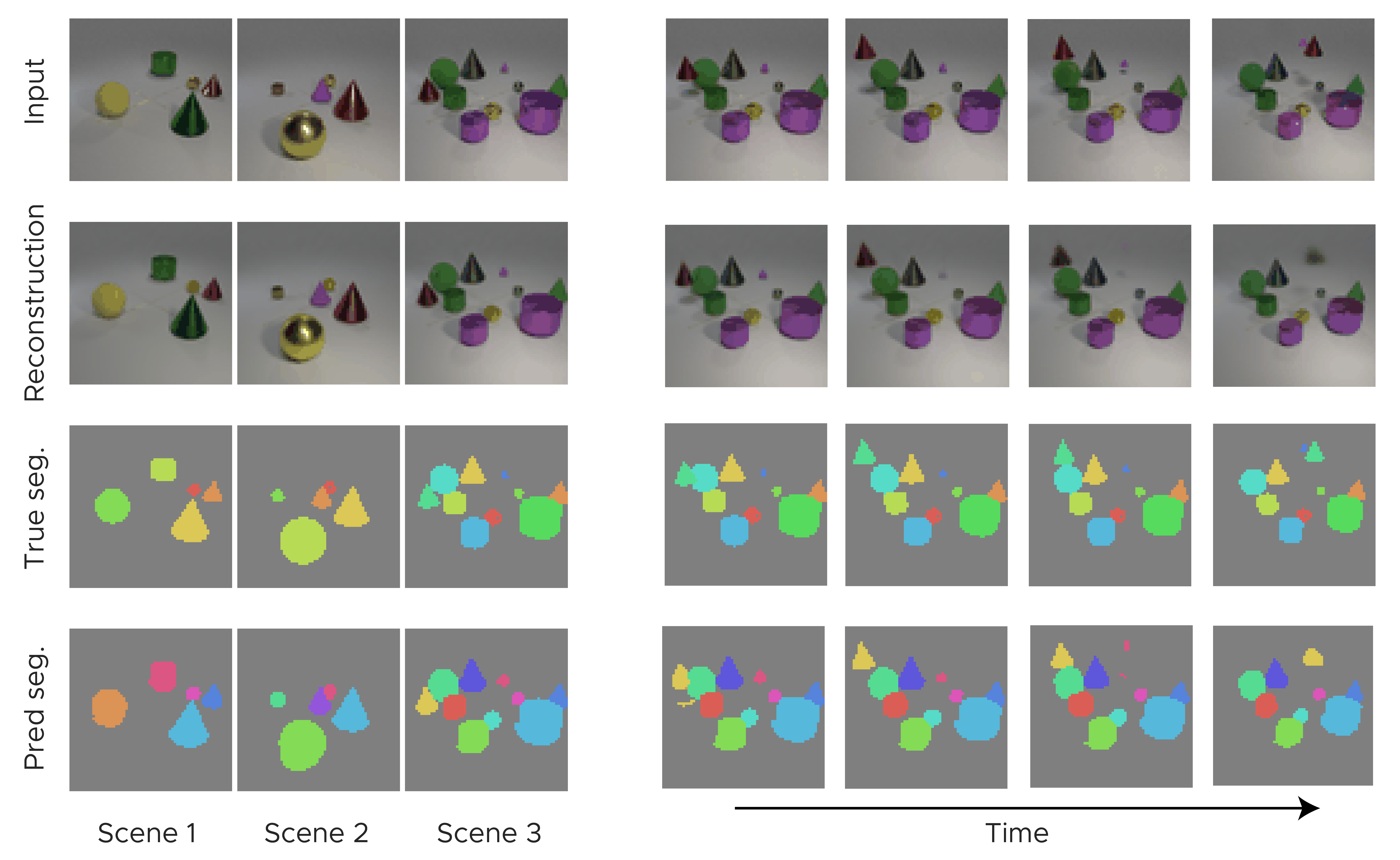}
  \caption{\textbf{Model performance for reconstructing and segmenting novel scenes.} (Left) Model performance across three scenes taken from the test set of CATER. The top two rows show model input and reconstruction of input, the bottom two rows show the true segmentation of objects within the scene and the predicted segmentation of the model. (Right) Reconstruction and segmentation performance across time for Scene 3. The top left object (red in the original input) moves up and to the left of the frame. The model is able to accurately segment and track the object across all 16 frames (4 intermediate frames shown here), albeit classifying the shadow as part of the object. 
}
  \label{fig4}
\end{figure*}

\subsection{Model performance on CATER and MOVi}

Lastly, we evaluated our neural network architecture on the Compositional Actions and TEmporal Reasoning (CATER) benchmark \cite{girdhar2019cater} as well as the MOVi-{A,B,C} datasets \cite{greff2022kubric}. These datasets consist of three to eleven randomly placed objects, with a fixed camera location but moving objects. We use the modified CATER dataset from \cite{kabra2021simone}, which includes segmentation masks for each object in order to explore the model's ability to perform object segmentation fully unsupervised. We use the same model architecture for higher-level cortices as described above, but replace the pre-trained visual representation with four convolutional layers, using a kernel size of four and a stride of 2. We additionally increase the number of units in the pixel-wise decoder from 128 to 512 and train the model for 300000 steps using a batch size of 1. As shown in Figure \ref{fig4}, our model is able to reconstruct the input frames and segment the objects on the CATER dataset, by using only a reconstruction loss. We compare our model against other unsupervised segmentation models like MONet \cite{burgess2019monet}, S-IODINE \cite{greff2019multi}, SIMONe \cite{kabra2021simone} and SAVi \cite{kipf2021conditional} (see Table \ref{tab1}). We achieve comparable or better performance to the best baseline models on CATER (FG-ARI: SAVi 0.928 $\pm$ 0.008; Ours 0.939 $\pm$ 0.013) while outperforming all unsupervised models (SAVi is trained on optical flow information). Based on visual examination of background segmentation in \cite{kabra2021simone}, we note that the full ARI score (taking into account both background and foreground) of our model likely outperforms many of the baseline models, which do not report the full score (Ours, FG-ARI: 0.939 $\pm$ 0.013, ARI: 0.825 $\pm$ 0.04). For the more challenging MOVi datasets, we further increase the number of units in the decoding layer to 1024 and train the model for 400000 steps using a batch size of 1. We observe a decline in performance with increasing scene complexity (FG-ARI, MOVi-A 0.790, MOVi-B 0.460, MOVi-C 0.318), while still outperforming SIMONe on all three datasets. Taken together, these comparisons show that our biologically inspired model is not only able to reconstruct images from our novel benchmark but shows comparable performance to state-of-the-art unsupervised scene segmentation models. 

\section{Discussion}

Here we explored the neural representations of an artificial neural network which was trained to perform allocentric topographical processing, similar to how the Four-Mountains-Test is used to predict the early onset of Alzheimer's disease. The model uses visual representations from V4 and IT and uses higher-level areas like the entorhinal and hippocampus to successfully differentiate between scenes from different viewpoints. We show that this biologically inspired model can discern between hundreds of scenes and generalize beyond its training set. Moreover, it is able to reconstruct the visual input through a factorized latent space \cite{kubilius2018cornet, kabra2021simone}, disentangling object from spatial information. 

Our model is able to perform novel view synthesis by imagining scenes from different viewpoints (4MT) or different moments in time (CATER \& MOVi) and can segment objects on par with recent state-of-the-art models. One shortcoming of this approach is the relatively small model size which likely prevents it from performing well on more challenging real-world datasets like MOVi-{C,D,E} or COCO. More powerful visual representations might help for these datasets, for which our visual cortex can be easily replaced with features from models that have a higher similarity to visual cortices \cite{schrimpf2020brain}. 

In the future, we want to further explore the difference in neural representations across the objective functions used. We observed that spatially modulated cells also arise in the model using a reconstruction loss, but only in the temporally averaged pathway (MEC) and only in a small subset of neurons. This likely arises from the use of LEC \& MEC as a bottleneck, forcing the model to retain scene information in order to fully reconstruct the image, which is not needed for the disentangling of the latent representation. 

{\small
\bibliographystyle{ieee_fullname}
\bibliography{main}
}

\setcounter{figure}{0}
\begin{figure*}
  \renewcommand{\thefigure}{S\arabic{figure}}
  \centering
  \includegraphics[width=\linewidth]{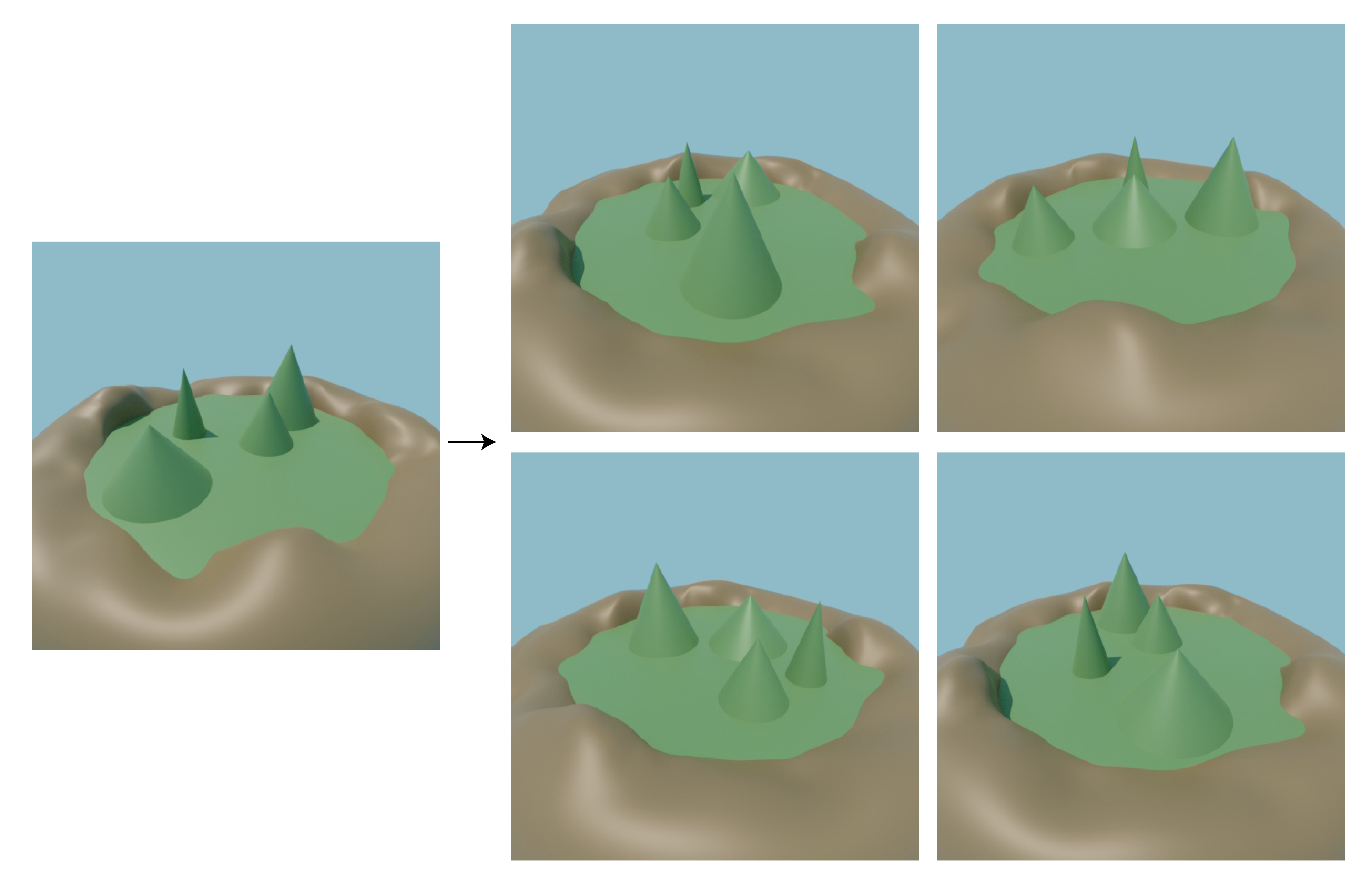}
  \caption{\textbf{Original task design, used to detect Alzheimer's disease} \\ In the 4-Mountains-Test \cite{hartley2007hippocampus}, the participant first sees the image on the left and after a two-second delay is shown the images on the right, from which they need to choose the image which shows the same scene but from a different viewpoint. In all other images, the allocentric position of the mountains is changed (distractor images). The correct scene in this example is the lower right image.}
  \label{fig_s1}
\end{figure*}

\begin{figure*}
  \renewcommand{\thefigure}{S\arabic{figure}}
  \centering
  \includegraphics[width=\linewidth]{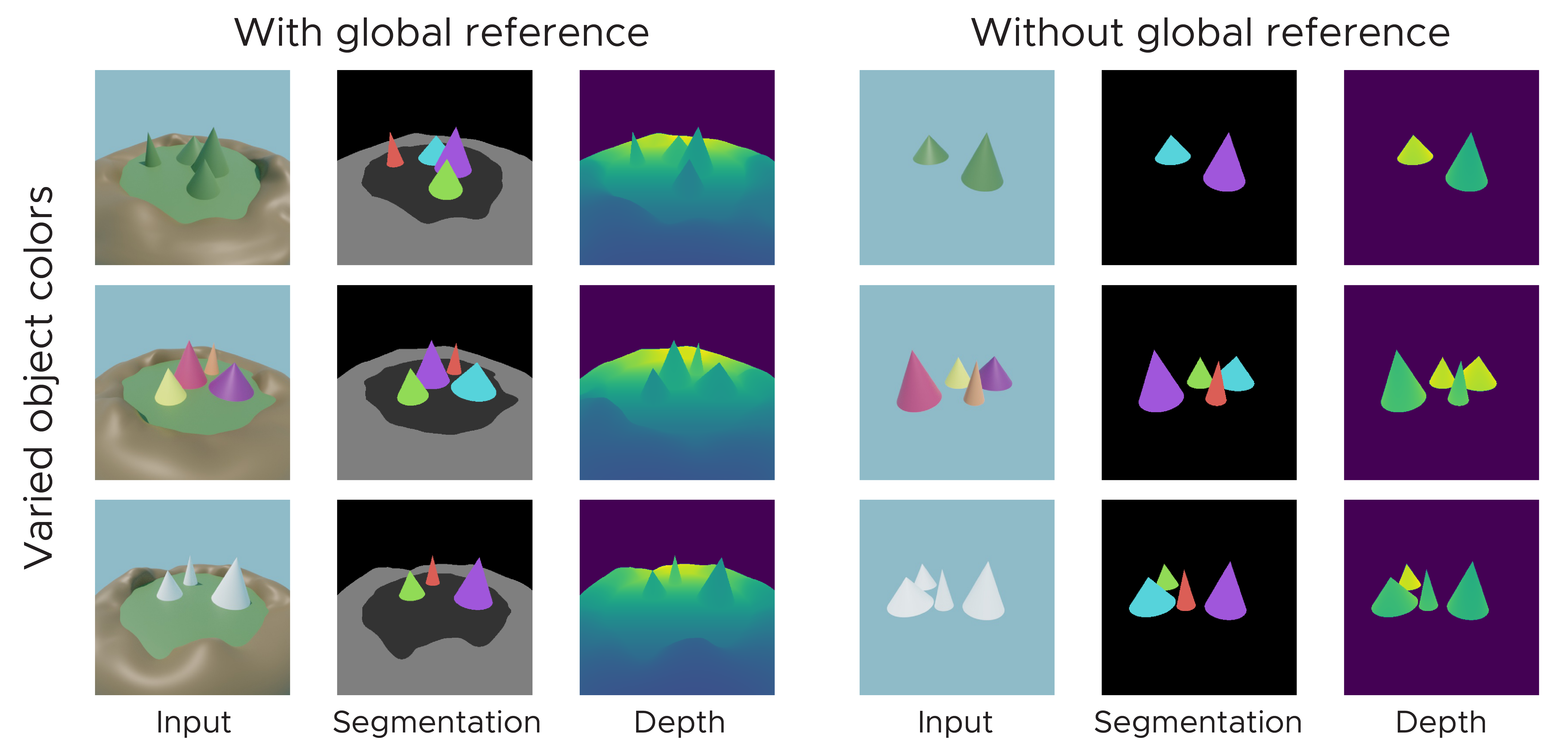}
  \caption{\textbf{Variations of allocentric scene perception (ASP) task} \\ We render six different variations of our dataset, by changing the colours of the object as well as the global reference. The colours are chosen to test model performance for segmenting objects with the same colour, objects which share the colour of the floor or are maximally different than the floor and objects which are already distinguishable by colour alone. We also vary the use of distal landmarks (global reference) to test model segmentation without additional cues regarding distances from boundaries in allocentric space. }
  \label{fig_s2}
\end{figure*}

\begin{figure*}
  \renewcommand{\thefigure}{S\arabic{figure}}
  \centering
  \includegraphics[width=\linewidth]{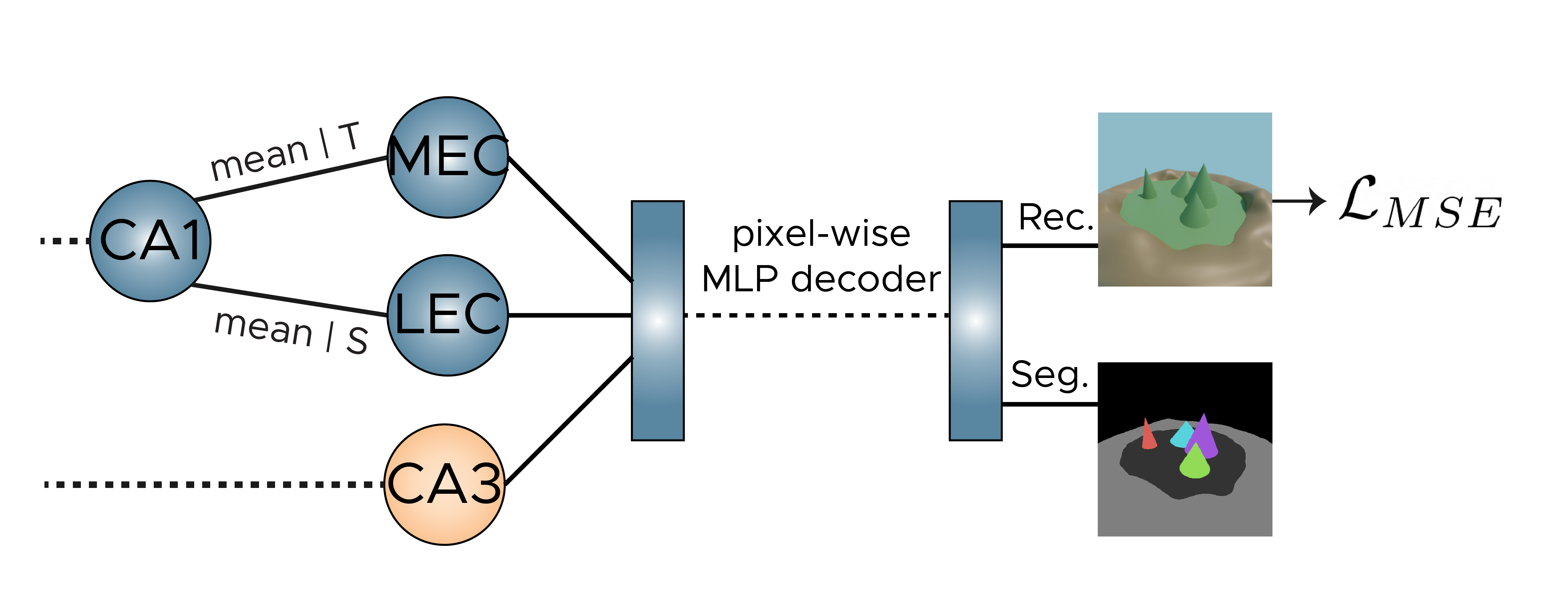}
  \caption{\textbf{Model architecture for reconstructing inputs} \\ When the model is tasked to reconstruct the input as well as segment objects, we use a pixel-wise decoder which takes input signals from MEC, LEC and CA3. MEC and LEC information is acquired by splitting the information into two pathways, one of which takes averaged time (T) information, thereby carrying information about space (MEC). In contrast, the other pathway takes averaged space (S) information, thereby retaining information about time (LEC). The MLP outputs four channels, where three are used for reconstructing the inputs, while the fourth is used for constructing segmentation masks. }
  \label{fig_s4}
\end{figure*}

\begin{figure*}
  \renewcommand{\thefigure}{S\arabic{figure}}
  \centering
  \includegraphics[width=\linewidth]{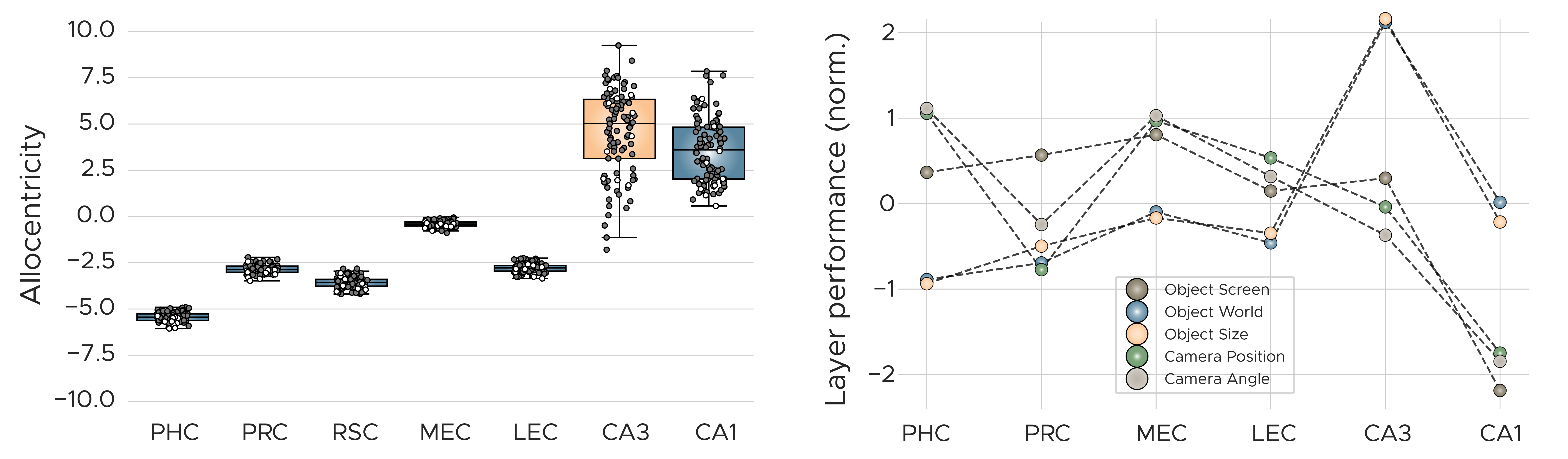}
  \caption{\textbf{Allocentricity score and layer readout} \\  (Left) We show the allocentricity score across layers, which measures the activity of neurons across different views within a scene, defined as the coefficient of variation of the activation of each artificial neuron across several images of the same scene. High values indicate that the neurons within this layer have similar activity profiles for images from the same scene independent of the viewpoint from where it was taken. Most world-centred responses are measured in late layers, indicating that hippocampal responses are similar for images from the same scene from different viewpoints. (Right) We use a linear readout across layers to investigate the information contained within them. Performance is normalized via z-scoring across layers. }
  \label{fig_s3}
\end{figure*}

\begin{figure*}
  \renewcommand{\thefigure}{S\arabic{figure}}
  \centering
  \includegraphics[width=\linewidth]{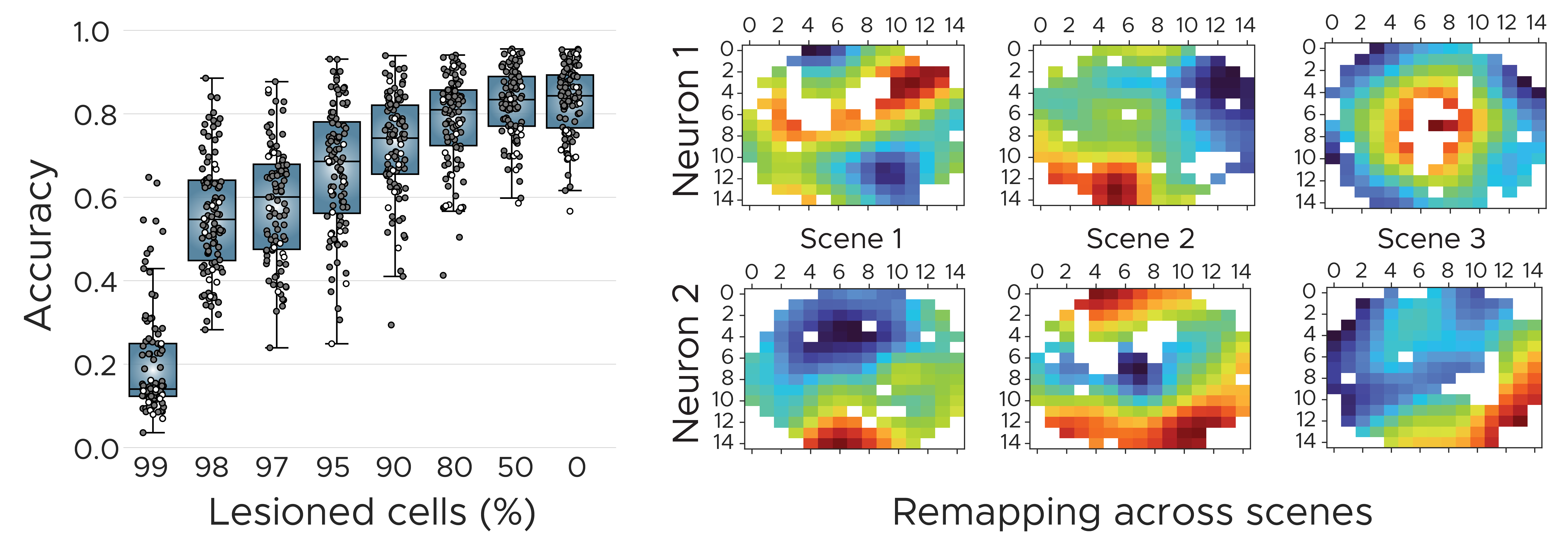}
  \caption{\textbf{Lesioning experiments and remapping across scenes} \\  (Left) Performance for differentiating scenes across the ratio of lesioned cells. (Right) Responses of two neurons across three different scenes, showing spatial selectivity remaps with scene identity. }
  \label{fig_s5}
\end{figure*}

\end{document}